\pdfoutput=1
\documentclass[12pt]{article}

\usepackage{amsmath,amssymb,latexsym}
\usepackage{bm}
\usepackage{graphicx}
\usepackage{fullpage}
\usepackage{hyperref}
\usepackage{lineno}

  \newcommand{\beginsupplement}{%
        \setcounter{table}{0}
        \renewcommand{\thetable}{S\arabic{table}}%
        \setcounter{figure}{0}
        \renewcommand{\thefigure}{S\arabic{figure}}%
         \setcounter{section}{0}
        \renewcommand{\thesection}{S\arabic{section}}
         \setcounter{equation}{0}
         \renewcommand{\theequation}{S\arabic{equation}}
     }

\newcommand{\figref}[2]{Fig.\ \ref{#1}#2}

\begin{document}

{\huge\bfseries \noindent{Element-centric clustering comparison unifies overlaps and hierarchy}}\\[0.25in]

{\Large
\noindent{Alexander J. Gates\textsuperscript{a,1}, Ian B. Wood\textsuperscript{b,c}, William P. Hetrick\textsuperscript{d}, Yong-Yeol Ahn\textsuperscript{b,c,e,1}}} \\[0.2in]
{\normalsize\textsuperscript{a}Department of Physics, Northeastern University. Boston, MA.\\
{\normalsize\textsuperscript{b}Department of Informatics, Indiana University. Bloomington, IN.\\
\textsuperscript{c}Center for Complex Networks and Systems Research, Indiana University. Bloomington, IN.\\
\textsuperscript{d}Department of Psychological and Brain Sciences, Indiana University. Bloomington, IN.\\
\textsuperscript{e}Program in Cognitive Science, Indiana University. Bloomington, IN.\\
\textsuperscript{1}To whom correspondence should be addressed. \\ E-mails: \href{mailto:a.gates@northeastern.edu}{a.gates@northeastern.edu}  and \href{yyahn@indiana.edu}{yyahn@indiana.edu}}

\section*{Abstract}
Clustering is one of the most universal approaches for understanding complex data.
A pivotal aspect of clustering analysis is quantitatively comparing clusterings; clustering comparison is the basis for many tasks such as clustering evaluation, consensus clustering, and tracking the temporal evolution of clusters.
In particular, the extrinsic evaluation of clustering methods requires comparing the uncovered clusterings to planted clusterings or known metadata.
Yet, as we demonstrate, existing clustering comparison measures have critical biases which undermine their usefulness, and no measure accommodates both overlapping and hierarchical clusterings.
Here we unify the comparison of disjoint, overlapping, and hierarchically structured clusterings by proposing a new element-centric framework: elements are compared based on the relationships induced by the cluster structure, as opposed to the traditional cluster-centric philosophy.
We demonstrate that, in contrast to standard clustering similarity measures, our framework does not suffer from critical biases and naturally provides unique insights into how the clusterings differ.
We illustrate the strengths of our framework by revealing new insights into the organization of clusters in two applications: the improved classification of schizophrenia based on the overlapping and hierarchical community structure of fMRI brain networks, and the disentanglement of various social homophily factors in Facebook social networks.
The universality of clustering suggests far-reaching impact of our framework throughout all areas of science.

\pagebreak

\section*{Introduction}

Clustering is one of the most basic and ubiquitous methods to analyze data \cite{Jain1999clustering, Fortunato2010communities}.
Traditionally, clustering is viewed as separating data elements into disjoint clusters of comparable sizes. 
Complications to this simplistic picture are becoming more prevalent, particularly following the rise of network science and nuanced clustering methods that reveal heterogeneous cluster size distributions~\cite{He2009imbalanceddata, Leskovec2009naturalcommunities}, overlaps~\cite{Palla2005overlapcomm, Ahn2010link, Gopalan2013overlapsbm, Yang2014overlapgroundtruth}, and hierarchical structure~\cite{Ravasz2002hiermetabolic, SalesPardo2007hiersystems, Delvenne2010stability, Zhang2014scalablemodularity}.
A growing consensus suggests that applying clustering is more about identifying appropriate techniques for the particular problem and properly interpreting the results, than developing a silver-bullet clustering method \cite{Kleinberg2002clusteringimpossiblity, Peel2017metadata}.

The most fundamental step towards understanding, evaluating, and leveraging identified clusterings is to quantitatively compare them.
Clustering comparison is the basis for clustering evaluation, consensus clustering, and tracking the temporal evolution of clusters, among many other tasks.
The proliferation of nuanced clustering methods presents new challenges for clustering comparison \cite{Meila2007compareclusteringsinfo, He2009imbalanceddata} and renders current methods susceptible to critical biases \cite{White1994nmibias, Pfitzner2009characterizingclusterings, He2009imbalanceddata, Vinh2010variantsMI, Zhang2015nmiproblems,  Gates2017impact}.
In addition to the consistent grouping of elements into clusters, similarity measures must account for many other aspects of clusterings, such as the number of clusters, the size distribution of those clusters, multiple element memberships when clusters overlap, and scaling relations between levels of hierarchical clusterings.

Despite the increasing prevalence of irregular cluster features, the effect of such structure on clustering similarity has received little attention.
Here we illustrate that the most popular clustering similarity measures are vulnerable to critical biases, calling the appropriateness of their general usage into question.
We also argue that these biases are maintained or exacerbated by extensions to accommodate overlapping or hierarchical clusterings \cite{Fowlkes1983hierarchicalcompare, Collins1988omega, Lancichinetti2009onmi, Perotti2015hierarchicalnmi}, suggesting that none of the existing frameworks for clustering similarity are adequate for comparing overlapping and hierarchically structured clusterings.

Here we propose a new \emph{element-centric} framework for clustering similarity that naturally incorporates overlaps and hierarchy.
In our approach, elements are compared based on the relationships induced by the cluster structure, in contrast to the traditional \emph{cluster-centric} philosophy.
As we will see, this change in perspective resolves many of the aforementioned difficulties and avoids the common biases induced by irregular cluster structure.

\section*{Bias in clustering comparisons}

Every clustering similarity measure must trade-off between variation in three primary characteristics of clusterings: the grouping of elements into clusters, the number of clusters, and the size distribution of those clusters \cite{Meila2005comparingclusteringsaxiom, Albatineh2006corrchance, Amigo2009clusteringcomparison, Pfitzner2009characterizingclusterings, Souto2012imbalancedclustering, Gates2017impact}. 
A failure to account for all three characteristics can result in a biased comparison in which clusterings with exaggerated features are favored over more intuitively similar clusterings.
Before exploring these trade-offs further, we offer an illustrative example by the comparisons between clustering pairs shown in \figref{fig:biasexample}{}.
Here we focus on three exemplary similarity measures---the normalized mutual information (NMI), Fowlkes-Mallows index (FM), and our element-centric similarity measure---and extend our discussion to a larger selection later.
In the first set of comparisons (\figref{fig:biasexample}{a}), we demonstrate a bias towards clusterings with heterogeneous cluster sizes: NMI and the element-centric similarity determine the middle clustering is more similar to the left clustering than the right clustering, yet FM concludes the opposite---the middle clustering is more similar to the right clustering than the left---as it is biased by the large cluster in the right clustering.
In the second set of comparisons (\figref{fig:biasexample}{b}), we illustrate a bias towards clusterings with more clusters: FM and the element-centric similarity determine the middle clustering is more similar to the left clustering than the right clustering, yet NMI concludes the opposite---the middle clustering is more similar to the right clustering than the left clustering---as it is biased by the number of clusters in the right clustering.

% 17.8cm, 11.4cm
%%%%%%%%%%%%%%%%%%%%%%
\begin{figure*}
\begin{center}
	\includegraphics[width = 12cm]{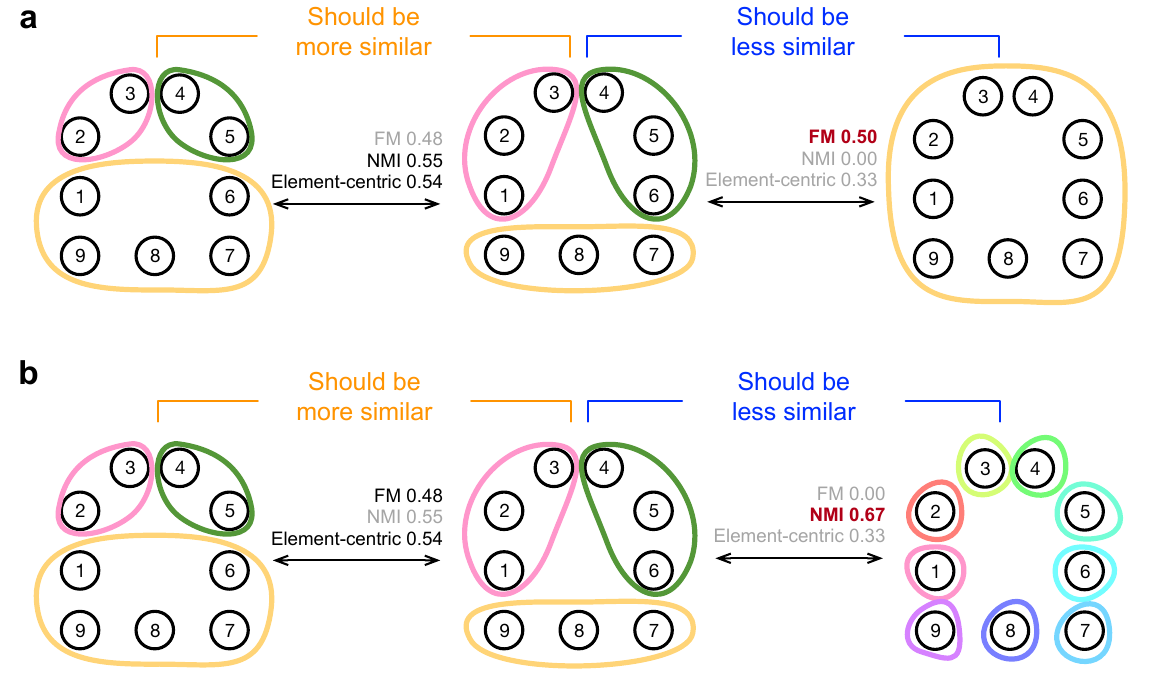}
	\caption{Two examples of counter-intuitive bias in clustering comparisons. Four clusterings are considered over 9 elements, and compared using the Fowlkes-Mallows index (FM), normalized mutual information (NMI), and our element-centric similarity measure. We argue that the comparison between the clusterings on the left is more similar than the comparison between clusterings on the right.  \textbf{a}, Both NMI and the element-centric similarity follow this intuition, but FM is biased towards large clusters and suggests the comparison on the right is more similar.  \textbf{b}, Both FM and the element-centric similarity follow this intuition, but NMI is biased towards many clusters and suggests the comparison on the right is more similar.}
 \label{fig:biasexample}
 \end{center}
\end{figure*}
%%%%%%%%%%%%%%%%%%%%%%%%

One approach to correct biases in clustering comparison is to consider clustering similarity in the context of a random ensemble of clusterings \cite{Hubert1985adjrand, Albatineh2006corrchance, Vinh2009nmicorrection, Vinh2010variantsMI, Albatineh2011clusteringchance, Romano2014standardized}.
Such a correction for chance uses the expected similarity of all pair-wise comparisons between clusterings specified by a random model to establish a baseline similarity value.
However, the correction for chance approach has severe drawbacks \cite{Gates2017impact}: (i) it is strongly dependent on the choice of random model assumed for the clusterings, which is often highly ambiguous, 
and (ii) no random model for overlapping or hierarchical clusterings has been suggested.

We introduce a simple set of synthetic clustering examples that illustrate the trade-offs between characteristics of clusterings.
In each case, we outline the desired behavior for a measure of clustering similarity based on the extensive discussion in the literature \cite{Fowlkes1983hierarchicalcompare, Hubert1985adjrand, White1994nmibias, Meila2005comparingclusteringsaxiom, Rosenberg2007vmeasure, Pfitzner2009characterizingclusterings, Amigo2009clusteringcomparison, Vinh2009nmicorrection, Souto2012imbalancedclustering, Amelio2015nmifair, Gates2017impact, Hanneke2018info}.
Our intuition is based on the use of clustering similarity in practice: similar clusterings should have a similar number of clusters, of similar sizes, and elements should have similar memberships.
Consider a typical case facing a practitioner of data science: we have three clustering methods M1, M2, and M3 such that M1 produces the clustering on the left of \figref{fig:biasexample}{a,b}, M2 produces the clustering on the top right of \figref{fig:biasexample}{a}, and M3 produces the clustering on the bottom right of \figref{fig:biasexample}{b}. 
Which method, M1, M2, or M3 performed best in recovering the ground-truth clustering in the middle of \figref{fig:biasexample}{a,b}?
The answer depends on the clustering similarity measure used.
Yet, to our best understanding, clustering M1 is the only clustering that reflects the number, sizes, and memberships of the ground-truth clustering.
While other intuitions are possible (i.e.\ that offered by information theory or the correction for chance as discussed further in the SI, Section S2.11), we argue that the intuition adapted here most accurately captures the use of clustering comparisons in the literature \cite{Lancichinetti2008benchmark}.

Since it is difficult to isolate changing cluster sizes or number of clusters from the grouping of elements into clusters, our examples consider the case of randomized element memberships; however, in practice, quantitative comparisons would make simultaneous trade-offs between all three aspects of clusterings.
Here we expand our focus to six exemplary similarity measures representing many of the most common measures from the literature---the Jaccard Index, Adjusted Rand Index (ARI)\cite{Hubert1985adjrand}, F measure, Fowlkes-Mallows index (FM)\cite{Fowlkes1983hierarchicalcompare}, percentage matching (PM), the normalized mutual information (NMI), overlapping normalized mutual information (ONMI)\cite{Lancichinetti2009onmi}, and our element-centric similarity measure.
We discuss another popular measure, the variation of information, in the SI, Section S2.10, due to its interpretation as a distance measure.
These four examples suggest that the most common clustering similarity measures are subject to critical biases which render them inappropriate for comparing generalized clusterings---only our element-centric similarity measure displays the intuitive behavior in all examples and does not suffer from the problem of matching (\figref{fig:similaritybias}{d}).

% 17.8cm, 11.4cm
%%%%%%%%%%%%%%%%%%%%%%
\begin{figure*}
\begin{center}
	\includegraphics[width = 15cm]{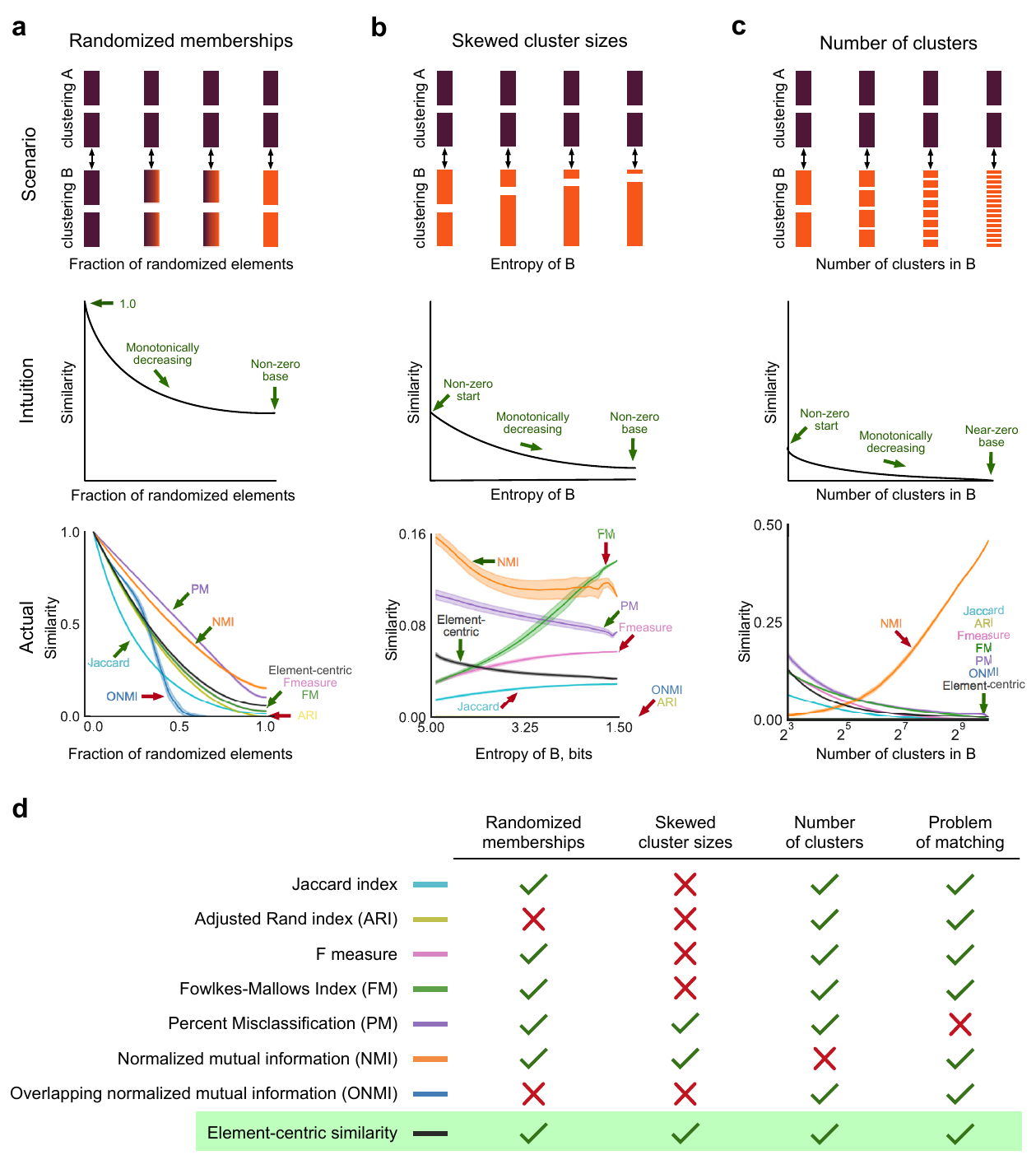}
	\caption{Element-centric similarity behaves intuitively in three clustering similarity scenarios while common clustering similarity measures exhibit counter-intuitive behaviors.  1,024 elements are assigned to clusters according to the following scenarios (\textbf{a-c}) and compared using the Jaccard index, adjusted Rand index, the F measure, percentage matching (PM), normalized mutual information (NMI), overlapping normalized mutual information (ONMI), and our element-centric similarity.  All results are averaged over 100 runs and error bars denote one standard deviation. \textbf{a}, A clustering with 32 non-overlapping and equal-sized clusters is compared to a randomized version of itself where elements are shuffled.  \textbf{b}, A clustering with 32 non-overlapping and equal-sized clusters is compared against clusterings with increasing cluster size skewness. \textbf{c}, A clustering with 8 non-overlapping and equal-sized clusters is compared against a clustering with $n$ non-overlapping, equal-sized clusters and randomized element memberships for different values of $n$.  \textbf{d}, Only our element-centric similarity measure follows the intuitive behavior in all three scenarios and does not suffer from the problem of matching.}
 \label{fig:similaritybias}
 \end{center}
\end{figure*}
%%%%%%%%%%%%%%%%%%%%%%%%

\subsection*{Bias in Randomized Membership}

In the first example, the consistent grouping of elements is tested by comparing a clustering of 1,024 elements into 32 equally sized clusters against itself after a fraction of element memberships have been shuffled between clusters (\figref{fig:similaritybias}{a}).
Intuition suggests that as the randomization increases, the similarity between the original clustering and the shuffled clustering should decrease from the maximum value ($1.0$ in all cases) to some non-zero value, reflecting the fact that the number and sizes of clusters are still identical. 
However, two measures reach zero, ignoring the similarity of the cluster size sequences.
The ONMI is particularly conservative, reporting no similarity at just over $50\%$ randomization; ONMI's surprising behavior highlights the difficulty of accommodating overlaps in a traditional similarity framework.

\subsection*{Bias in Skewed Cluster Sizes}

The second example explores the bias favoring skewed cluster size sequences through a preferential attachment shuffling scheme (\figref{fig:similaritybias}{b}).
Starting from the same initial clustering of 1,024 elements into 32 equally sized clusters, we randomize all element memberships.
The algorithm then proceeds to uniformly select a random element and reassign it to a new cluster based on the current sizes of those clusters.
This procedure is run for a total of $5\times10^6$ steps, with a comparison to the original clustering performed every 500 steps.
We argue that the desired clustering comparison behavior should reflect the cluster size differences, and that a decrease in the entropy of the cluster size sequence (reflecting an increase in cluster size heterogeneity) is reflected by the two clusterings becoming less similar.
However, we now see three distinct types of behaviors exhibited by the clustering similarity measures.
The NMI and our element-centric similarity measure exhibit the intuitive behavior and decrease as the clustering entropy decreases.
The ONMI and ARI maintain a zero similarity for all comparisons regardless of the clustering entropy.
Finally, the F measure and Jaccard index increase as the entropy decreases: They cannot account for the differences in the cluster size distribution.
This increase is a consequence of their formulation in terms of the correctly co-assigned element pairs while disregarding the incorrectly co-assigned element pairs.

\subsection*{Bias in The Number of Clusters}

Third, we investigate a scenario where the number and sizes of clusters in two clusterings diverge (\figref{fig:similaritybias}{c}).
Here we compare an initial clustering of 1,024 elements into 8 equally sized clusters against a second clustering generated by randomly assigning the elements to $c$ regularly sized clusters, where $c$ is the control parameter for the scenario.
Hence, one clustering remains the same size, while the other has $c$ regularly sized clusters.
We see two distinctly different behaviors of the clustering similarity measures: the Jaccard index, F measure, ONMI, ARI and our element-centric similarity measure all follow our intuition and decrease with increasing $c$, while NMI increases with increasing $c$.  
The increasing behavior for NMI can be attributed to the aforementioned information-theoretic bias towards comparisons with more clusters \cite{White1994nmibias, Zhang2015nmiproblems, Amelio2015nmifair, Amelio2016nmicloseness, Gates2017impact}, and counters the large body of established literature controlling for the number of clusters in a clustering solution \cite{Tibshirani2001gapstat, Newman2016numcomm}.
This bias makes NMI a particularly troubling measure for hierarchical clusterings where we expect the number of clusters to vary over several orders of magnitude.

\subsection*{The Problem of Matching}

Finally, we recount one of the oldest biases discussed in the literature, the problem of matching \cite{Meila2003comparingvi, Meila2007compareclusteringsinfo, Rezaei2016setmatch}.
The problem of matching is a symptom of all set-matching methods which identify a ``best match'' for each cluster. 
As a result, the measures completely ignore what happens to elements in the ``unmatched'' part of each cluster. 
For example, suppose $\mathcal{A}$ is a clustering with $K$ equal-sized clusters over $N$ elements, with $N\gg K$, and clustering $\mathcal{B}$ is obtained from $\mathcal{A}$ by moving a small fraction of the elements in each cluster $\mathcal{A}_k$ to the cluster $\mathcal{A}_{k+1\mod k}$. 
Likewise, the clustering $\mathcal{C}$ is obtained from $\mathcal{A}$ by reassigning the same fraction of the elements in each $\mathcal{A}_k$ evenly between the other clusters. 
In this case, measures suffering from the problem of matching would say the similarity between $\mathcal{A}$ and $\mathcal{B}$ is equal to the similarity between $\mathcal{A}$ and $\mathcal{C}$, contradicting the intuition that $\mathcal{A}$ is more similar to $\mathcal{B}$ than $\mathcal{C}$. 
For the measures considered here, only the percentage matching similarity measure suffers from the problem of matching. 
Despite this issue, it is important to note that the percentage matching has been used both in practice and in theory, typically when the clusterings are assumed to be relatively similar. 

\subsection*{Consequence for Extensions to Overlapping and Hierarchical Structure}
The three examples discussed in this section illustrate biases in the case of disjoint clusterings without hierarchy.  
Despite the increasing prevalence of overlapping and hierarchical structured clusterings, there is a lack of intuition for the trade-offs encountered by clustering similarity measures in the presence of such structure.
However, exploring the behavior of similarity measures when comparing partitions reveals useful insights into how these measures behave when comparing other clustering structures.
The presence of overlaps can exaggerate the heterogeneity in cluster sizes \cite{Yang2014overlapgroundtruth}, especially if one considers each overlap region as a separate cluster (i.e.\ as considered by the Omega index).
Since hierarchical clusterings reflect cluster structure over many scales, the sizes of these clusters typically vary by orders of magnitude; for example, the benchmark models typically used to capture hierarchical structure in networks are full $k$-ary trees, and thus the number of clusters grows exponentially in the number of levels \cite{Ravasz2002hiermetabolic, Lancichinetti2009onmi}.

All but one of the similarity measures for overlapping or hierarchical clusterings simplifies to one of the cases we have studied: the Omega index is equivalent to the adjusted Rand index for partitions \cite{Collins1988omega}, hierarchical mutual information reduces to NMI on partitions \cite{Perotti2015hierarchicalnmi}, and the Fowlkes-Mallows analysis of dendrograms considers each cut of the dendrogram independently, thus producing a curve of comparisons between partitions \cite{Fowlkes1983hierarchicalcompare}.
The overlapping normalized mutual information (ONMI) is the only measure which does not reduce to another measure on partitions \cite{Lancichinetti2009onmi}, yet we have demonstrated that it has particularly unintuitive behavior in our examples.
In sum, all existing measures for overlapping or hierarchical clusterings either inherit critical biases from their simpler counterparts on flat-partitions, or are inadequate for handling overlapping and hierarchical clusterings.

\section*{Element-centric clustering comparisons}

Our element-centric clustering similarity approach captures cluster-induced relationships between the elements through the \emph{cluster affiliation graph}, a bipartite graph where one vertex set corresponds to the original elements and the other corresponds to the clusters.
Specifically, a cluster affiliation graph is constructed for a clustering $\mathcal{C}$ of labeled elements $V = \{v_1, \ldots, v_N\}$ as a bipartite graph $\mathcal{B}(V \cup C,\mathcal{R})$ where one vertex set corresponds to the original elements $V$ and the other vertex set corresponds to the cluster set $C$.
An undirected edge $a_{i\beta}\in\mathcal{R}\subset V\times C$ is placed between element $v_i\in V$ and cluster $c_\beta\in C$ if $v_i\in c_\beta$, i.e.\ the element is a member of the cluster.  
Notice that an element's membership in multiple overlapping clusters can be directly incorporated with multiple edges in the cluster affiliation graph.
For hierarchically structured clusterings, each cluster $c_\beta \in C$ is assigned a hierarchical level $l_\beta\in[0,1]$ by re-scaling the hierarchy's acyclic graph (dendrogram) according to the maximum path length from the roots~\cite{Czegel2015rwhierarchy}.
The weight of the cluster affiliation edge is given by the hierarchy weighting function $h(l_\beta)$:
\begin{linenomath*}
    \begin{equation}
        h(l_\beta) = e^{rl_\beta},
    \end{equation}
\end{linenomath*}
where $r$ is a scaling parameter that determines the relative importance of membership at different levels of the hierarchy (further discussed below).

The cluster affiliation graph is then projected onto the element vertices to produce the \emph{cluster-induced element graph}, which is a weighted, directed graph that summarizes the inter-element relationships induced by common cluster memberships \cite{Zhou2007projection} (see \figref{fig:workflow}{c}).
In the cluster-induced element graph, with weighted adjacency matrix $\bm{W}$, each edge $w_{ij}$ between elements $v_i$ and $v_j$ has weight:
\begin{linenomath*}
    \begin{equation}
        w_{ij} = \sum_{\gamma} \frac{a_{i\gamma}a_{j\gamma}}{\sum_\kappa a_{i\kappa}\sum_ma_{m\gamma}} ,
    \end{equation}
\end{linenomath*}
where $a_{i\gamma}$ are the entries of the $N\times K$ bipartite adjacency matrix $\mathbb{A}$ for the cluster affiliation graph.  

%%%%%%%%%%%%%%%%%%%%%%
% Figure with CSTG
\begin{figure*}
\begin{center}
	\includegraphics[width = 10cm]{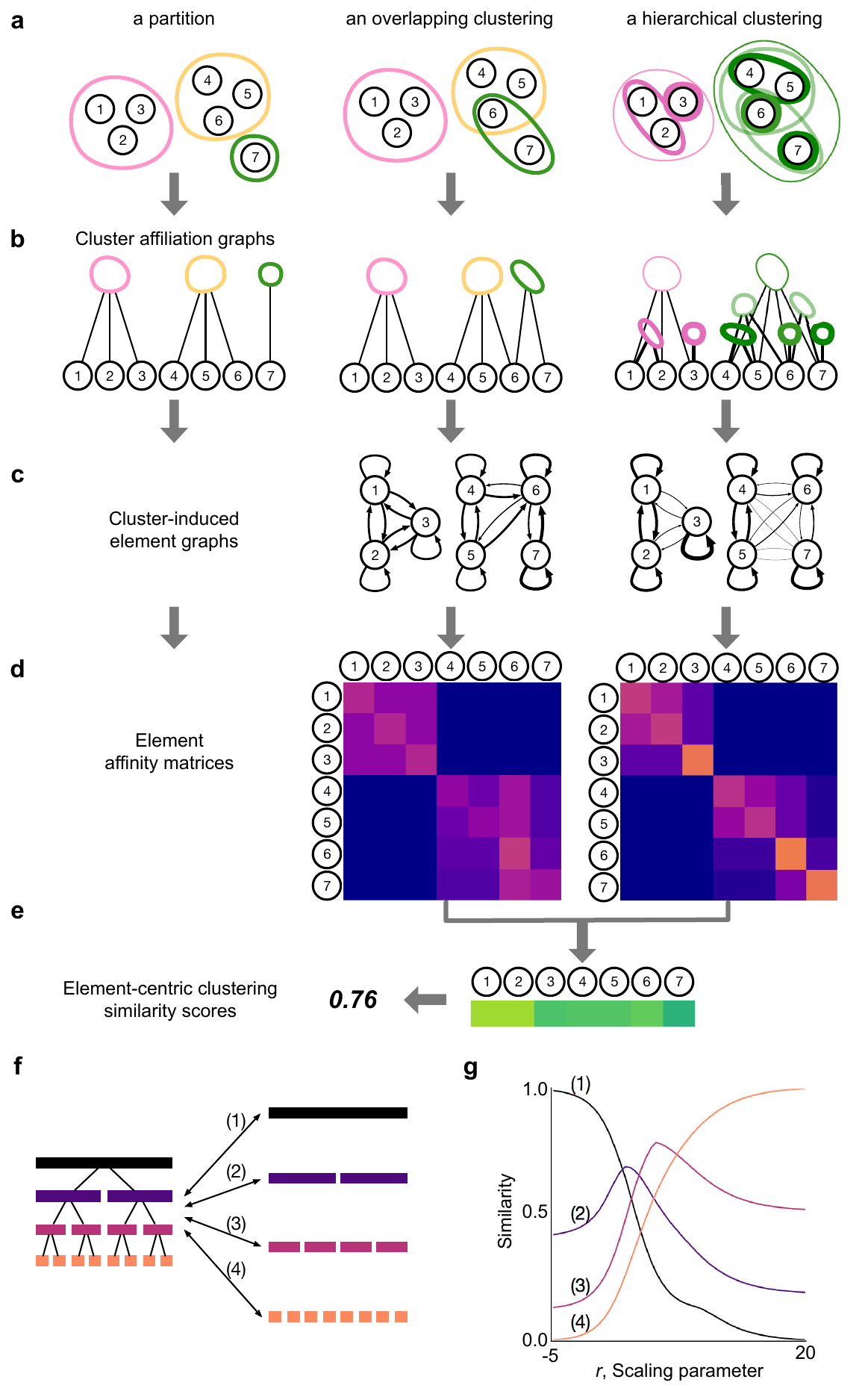}
	\caption{The element-centric perspective naturally incorporates overlaps and hierarchy.  \textbf{a}, Three examples of clusterings:  a partition, a clustering with overlap, and a clustering with both overlapping and hierarchical structure.  \textbf{b}, Cluster affiliation graphs derived from the overlapping and hierarchical clusterings.  \textbf{c}, Cluster-induced element graphs found by projecting the cluster affiliation graphs in \textbf{b} to the element vertices.  \textbf{d}, The element-affinity matrices found as the personalized pagerank equilibrium distribution. \textbf{e}, The corrected L1 metric distance between each affinity distribution in \textbf{d} gives an element-wise similarity between clusterings, the average element-wise similarity provides the final element-centric clustering similarity score. \textbf{f}, A binary hierarchical clustering is compared to each of its individual levels. \textbf{g}, The hierarchical scaling parameter for element-centric similarity acts as a ``zooming lens'', refocusing the similarity to different levels (1-4) of the hierarchical comparison in \textbf{f}.}
	\label{fig:workflow}
	\end{center}
\end{figure*}
%%%%%%%%%%%%%%%%%%%%%%%%

The traditional notion of pair-wise co-occurrence in a cluster is now captured by the (binary) presence of an edge in the cluster-induced element graph.
However, the focus on element \emph{pairs} misses high-order relations (triplets, quadruplets, etc.), which are useful for characterizing cluster structure \cite{Hubert1985adjrand}.
Such high-order co-occurrences can be captured through the presence of paths in the cluster-induced element graph.
The weight of the path accounts for the relative importance of elements in the presence of overlapping and hierarchical cluster structures.
Here, we incorporate every possible path between elements obtaining the equilibrium distribution for a personalized diffusion process on the graph (often called ``personalized pagerank'' or ``random walk with restart'') \cite{Haveliwala2003topicpagerank, Tong2006rwr, Kloumann2016blockppr}.
Given a cluster-induced element graph with weighted adjacency matrix $\bm{W}$, the personalized PageRank (PPR) affinity from element $v_i$ to all elements $v_j$ is found as the stationary distribution of a diffusion process with restart probability $1.0 - \alpha$ to $v_i$ which takes the form: 
\begin{linenomath*}
    \begin{equation}
        	\bm{p}_{i} = (1.0 - \alpha) \bm{v}_i + \alpha \bm{p}_{i}\bm{W},
    \end{equation}
\end{linenomath*}
where $\bm{v}_i$ is an $N$-vector with $1$ in the $i$th entry, and $0$ otherwise.
The value of $\alpha$ controls the influence of overlapping clusters and hierarchical clusters with shared lineages; here we use $\alpha = 0.90$.

In general, for large data sets and clusterings with many overlapping and hierarchical clusters, the calculation of personalized pagerank can be a computationally expensive process.
However, there are some computational simplifications that can be made.
First, the personalized PageRank affinity of partitions (disjoint clusterings) can be analytically solved---the affinity value for each co-clustered element pair is a linear function of the inverse cluster size $|c_\beta|$, and $0$ otherwise:
\begin{linenomath*}
    \begin{equation}
        p_{ij} = \left(\alpha/|c_\beta| + (1-\delta_{ij})(1-\alpha)\right)\delta_{\beta\gamma},
    \end{equation}
\end{linenomath*}
where $\delta$ is the Kronecker delta function, element $v_i$ is in cluster $c_\gamma$, and element $v_j$ is in cluster $c_\beta$.
Second, when several elements share exactly the same cluster memberships, their resulting personalized pagerank affinity vectors are related by simple permutations; therefore, the personalized pagerank affinity vector need only be calculated once for each common cluster membership set.
Third, due to the utility of personalized pagerank for recommendation systems, there have been many algorithms for the approximation of personalized pagerank \cite{Lofgren2014fastppr, Gleich2016seededpagerank}.
The worst-case computational complexity of element-centric similaritywill only occur for highly overlapping and deeply hierarchical clusterings, which were previously incomparable using traditional clustering similarity methods.

The element-wise similarity of an element $v_i$ in two clusterings $\mathcal{A}$ and $\mathcal{B}$ is found by comparing the stationary probability distributions $\bm{p}_i^{\mathcal{A}}$ and $\bm{p}_i^{\mathcal{B}}$ induced by the PPR processes on the two cluster-induced element graphs.  Here, we use the normalized L1 metric for probability distributions corrected to account for the PPR process:
\begin{linenomath*}
    \begin{equation}
        S_i(\mathcal{A}, \mathcal{B})= 1.0 - L1_\alpha(\bm{p}_{i}^{\mathcal{A}}, \bm{p}_{i}^{\mathcal{B}}) = 1.0 - \frac{1}{2\alpha}\sum_{j=1}^N |p_{ij}^{\mathcal{A}} - p_{ij}^{\mathcal{B}}| ,
    \end{equation}
\end{linenomath*}
The L1 metric was chosen because it is invariant to the magnitude of the probability values, i.e.\ it treats all cluster sizes equally.
Other popular probability metrics (Hellinger, Euclidean, etc.) extenuates the differences in small or large probability values, hence they would depend on the cluster sizes.
The final element-centric similarity score $S(\mathcal{A}, \mathcal{B})$ of two clusterings $\mathcal{A}, \mathcal{B}$ is the average of the element-wise similarities:
\begin{linenomath*}
    \begin{equation}
        S(\mathcal{A}, \mathcal{B}) = \frac{1}{N} \sum_{i=1}^N S_i(\mathcal{A}, \mathcal{B}).
    \end{equation}
\end{linenomath*}
A full implementation of the element-centric clustering similarity, and all other clustering similarity measures discussed here, is provided in the CluSim python package \cite{Gates2018clusim}.
As illustrated in \figref{fig:workflow}, our element-centric framework unifies disjoint, overlapping, and hierarchical clustering comparison in a single framework.

\section*{Interpretations of Element-Centric Similarity}

\subsection*{Cluster Affiliation Graph and Cluster-induced Element Graph}
The cluster affiliation graph provides a convenient representation of element membership in multiple clusters at different scales of the hierarchy.
Unweighted variants of the affiliation graph are common approaches to study the relationship between labels and data in network science~\cite{Zhou2007projection, Yang2012ctag}.
Our weighted extension reflects the varying importance of membership at different scales of the hierarchy.

The element-centric philosophy suggests a focus on common memberships between data elements induced by the cluster structure, rather than overlaps between clusters induced by elements (as suggested by the cluster-centric philosophy).
The cluster-induced element graph captures these relationships by integrating over all shared cluster memberships through the projection of the cluster affiliation graph onto the element nodes.
This projection has three important features.
First, the induced relationship between two elements is normalized by the size of the cluster capturing the fact that co-occurrence in larger clusters implies less direct influence between elements than co-occurrence in smaller clusters.
Second, the weight for each element is normalized by the sum over all of its cluster memberships reflecting the idea that membership in many clusters reduces the relative influence from any one of the clusters.
Third, in the presence of overlap or hierarchy, the weights in the cluster-induced element graph can be asymmetric (i.e. $w_{ij}\neq w_{ji}$) arising from the fact that multiple cluster affiliations will change the respective local neighborhoods of individual elements.
Note that our normalization for the edge-weights in the cluster-induced element graph is equivalent to the landing probability of a two-step random walk on the cluster affiliation graph from element $v_i$ to element $v_j$.

%An open question of interest would be to explore the structure of the cluster affiliation graph when projected onto the cluster nodes.  

\subsection*{Element-wise Scores}
Beyond naturally accommodating generalized clusterings, our element-centric similarity can provide detailed insights into how two clusterings differ because the similarity is calculated at the level of individual elements. 
Specifically, the individual element-wise scores $S_i(\mathcal{A}, \mathcal{B})$ directly measure how similar the clusterings appear from the perspective of each element.
The distribution of element-wise similarity scores can also provide insight into how the clusterings differ.
For example, the ranked-distribution of element-wise scores reflects the differences in cluster structure: a flat distribution occurs when all elements have the same similarity score, suggesting that the clusterings differ equally across all elements; a skewed distribution occurs when some elements have much higher or lower similarity than the rest, suggesting that the clusterings are distinguished by a subset of elements.

%
%The element-wise \emph{agreement} is revealed by the average of these element-wise scores over comparisons between uncovered clusterings and a reference clustering (SI, section S3.6).  %\ref{sec:agreement}).
%
%The element-wise scores can also be averaged over all pair-wise comparisons within the set of uncovered clusterings, revealing the \emph{frustrated} elements that cannot be consistently clustered.
%

\subsection*{Average agreement and frustration}
\label{sec:agreement}
Our element-centric  similarity measure also reveals the consistency of element groupings within an arbitrary set of clusterings.
The average \emph{agreement} between a reference clustering and a set of clusterings measures the regular grouping of elements with respect to a reference clustering.
Specifically, given a clustering $\mathcal{G}$ and a set of clusterings $\bm{R} = \{\mathcal{R}_1,\ldots, \mathcal{R}_T\}$, the element-wise average agreement for element $v_i$ is evaluated as:
\begin{linenomath*}
    \begin{equation}
        \frac{1}{T} \sum_{j=1}^T S_i(\mathcal{G}, \mathcal{R}_j).
    \end{equation}
\end{linenomath*}
The \emph{frustration} within a set of clusterings reflects the consistency with which elements are grouped by the clusterings.
For the set of clusterings $\bm{R} = \{\mathcal{R}_1,\ldots, \mathcal{R}_T\}$, the element-wise frustration for element $v_i$ is given by:
\begin{linenomath*}
    \begin{equation}
        \frac{1}{\binom{T}{2}} \sum_{j=2}^T \sum_{k=1}^{j-1} S_i(\mathcal{R}_k, \mathcal{R}_j).
    \end{equation}
\end{linenomath*}

\subsection*{Interpretation of Overlap}
The element-centric framework naturally incorporates the multiple memberships that occur in overlapping clusterings.
First, as discussed above, element membership in multiple clusters is directly captured by multiple edges in the cluster affiliation graph, and is propagated into asymmetric weights in the cluster-induced element graph.
Second, the integration over local paths through the personalized PageRank process means that the presence of multiple memberships for elements is not isolated to the overlapping elements, but propagates throughout the clusters which overlap.  
This is because shared elements introduce additional information into the system, namely that the clusters share common features.  
Essentially, in the absence of any other information, if two clusters overlap (share some elements), then their elements should be more similar compared with the case where the clusters are disjoint.

For example, let us simplify our discussion to talk about counting element triplets in the overlapping clustering from \figref{fig:workflow}{a}.  
When no overlaps are present, it is simple to declare whether all three elements co-occur within the same cluster or not; so elements 1,2,3 all co-occur in the pink cluster, but elements 1,2,4 do not.
However, in the presence of overlap, additional decisions must be made.  
Consider elements 4,6,7.  Elements 4 and 6 co-occur in the same yellow cluster, and elements 6 and 7 co-occur in the same green cluster, but how should one determine if the elements 4 and 7 co-occur?  
Clearly, this triplet has important information. 
Indeed, it specifically defines what the ``overlap'' means for this clustering.  
Thus, the element-centric similarity measure does not disregard this triplet, but retains it with a reduced weight determined by the $\alpha$ parameter. 
Namely, the triplet 4,6,7 co-occurs less strongly than the triplet without overlap 1,2,3.

In contrast, the omega index counts element co-occurrences very conservatively and states that elements 4 and 7 do not co-occur.  
It continues to make the distinction that 4 and 6 didn’t co-occur either because 4 doesn’t have the exact same memberships as 6.  
Thus it throws away valuable information about the cluster structure.

\subsection*{Interpretation of Hierarchy}

Our element-centric framework is flexible and allows natural choices to accommodate alternative interpretations of hierarchy.
For example, our choice of hierarchical weighting function and the scaling parameter, $r$, reflects a continuum in the hierarchy (\figref{fig:workflow}{g}): 
lower $r$ emphasizes higher levels and reflects a divisive hierarchy, in which lower levels of the dendrogram are treated as refinements of the higher levels, while larger $r$ puts emphasis on lower levels and reflects an agglomerative hierarchy, in which higher levels of the dendrogram are seen as a coarsening of the lower level cluster structure.
Other interpretations of hierarchy can be implemented by changing the specific hierarchical weighting function; for example, constant function ($r=0$ above) collapses the hierarchy into an overlapping clustering with each cluster weighted equally.

\subsection*{Relation to Other Similarity Measures}
Our choice of L1 comparisons between personalized pagerank distributions was based on a principled extension of element co-occurrence.
This choice can be replaced by another measure of graph similarity or probability metric with an alternative intuition of the trade-offs associated with clustering similarity \cite{Blondel2004graphsim}.
Indeed, several common clustering similarity measures can be recovered by adapting other choices of graph similarity; all pair-counting measures can be recovered from graph set operations between cluster-induced element graphs from disjoint clusterings.  
The Rand index, in particular, is recovered by applying the graph-edit distance between the two cluster-induced element graphs from disjoint clusterings.

 %%%%%%%%%%%%%%%%%%%%%%%%%%%%%%
% Element-wise Measures of Similarity
%%%%%%%%%%%%%%%%%%%%%%%%%%%%%%
\section*{Applications}
\subsection*{Element-centric comparisons reveal insights into how K-means clusterings differ}
Beyond serving as a global measure of clustering similarity, our element-centric similarity also provides detailed insights into how clusterings differ, in contrast to other measures.
Consider an illustrative example from K-means clustering shown in \figref{fig:elementwise}{a} and detailed in the SI, Section S3.1;
$19$ clusters were randomly placed in a square with a randomly selected arrangement (Gaussian blob, anisotropic blob, circle, or spiral) and size.
K-means has difficulty when the predefined clusters overlap or are circularly arranged \cite{Jain2010kmeans}.
This difficulty can be explicitly quantified by calculating the average element-wise similarity between the predefined clustering and $100$ uncovered clusterings (\figref{fig:elementwise}{b}).
The element-wise frustration, found by averaging over all pair-wise comparisons between the $100$ uncovered clusterings, reveals data points that are consistently grouped into similar clusters or are assigned to drastically different clusters (\figref{fig:elementwise}{c}).
The combination of similarity and frustration identifies specific elements which are consistently grouped into an incorrect cluster (\figref{fig:elementwise}{b,c}: high error, low frustration), or those elements which K-means cannot consistently decide on a grouping (\figref{fig:elementwise}{b,c}: low error, high frustration).

%%%%%%%%%%%%%%%%%%%%%%
% Figure for Element-wise comparison
\begin{figure*}
\begin{center}
	\includegraphics[width = 14cm]{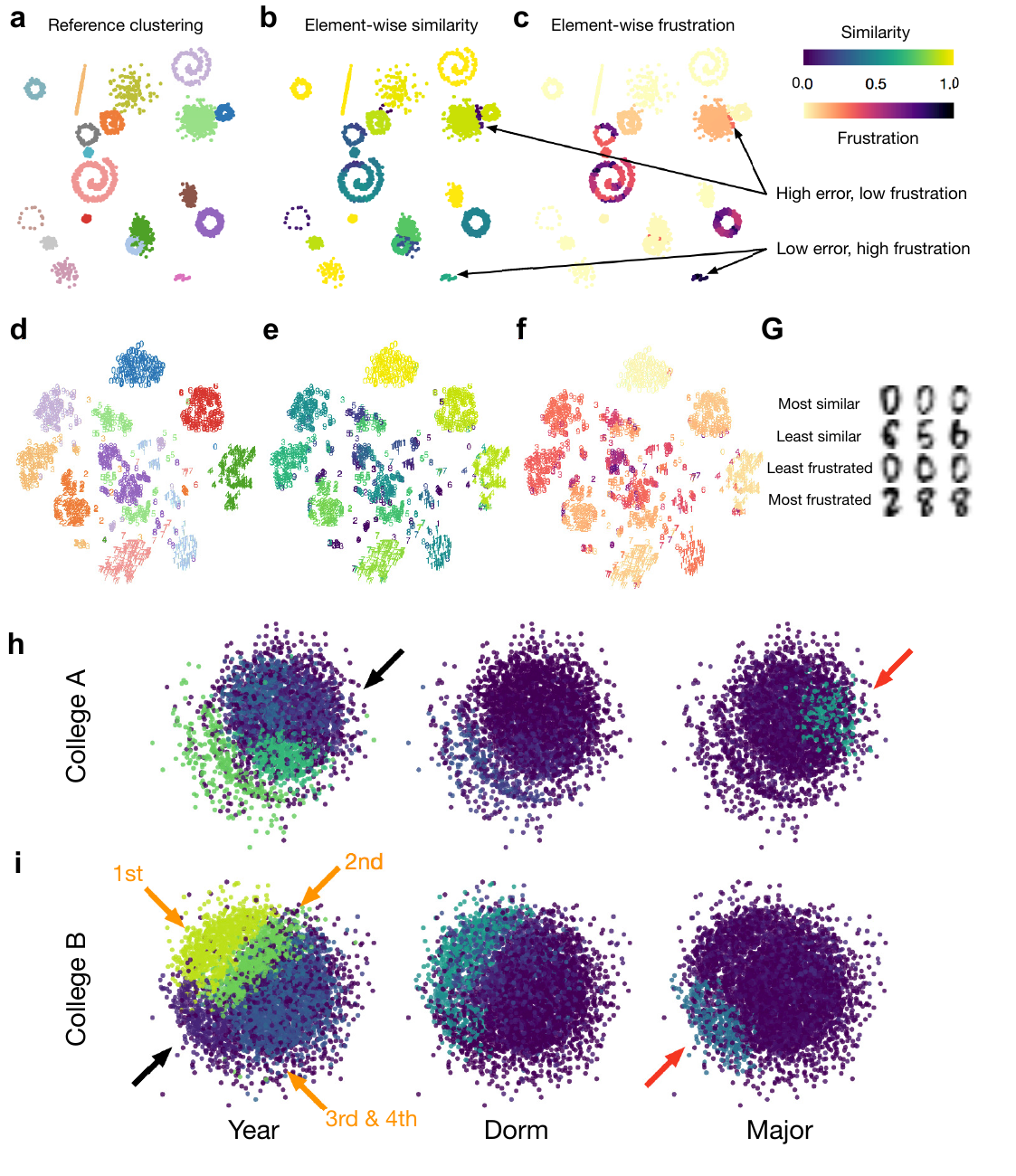}
	\caption{Element-wise clustering similarity reveals insights into how clusterings differ. \textbf{a-c}, A K-means clustering example. \textbf{a}, The planted clustering. \textbf{b}, The average element-wise similarity between the planted clustering and $100$ K-means clusterings. \textbf{c}, The average element-wise similarity between $100$ K-means clusterings.  \textbf{d-g}, A handwriting classification example. \textbf{d}, The labeled handwritten digit data projected using t-SNE dimensionality reduction for visualization. \textbf{e}, The average element-wise similarity between the labels and $100$ K-means clusterings. \textbf{f}, The average element-wise similarity between $100$ K-means clusterings. \textbf{g}, Exemplar digits that are consistently grouped as in the ground-truth clustering, consistently clustered differently from the ground-truth clustering, least frustrated, and most frustrated. \textbf{h,i}, Facebook friendship networks for \textbf{h} College A and \textbf{i} College B.  The element-wise similarity between user affiliation to school year, dorm, and major compared to Newman's modularity optimized by the Louvain method demonstrates that social networks can be organized by a convolution of different attributes (black vs red arrows).  The similarity to school year attenuates with student's status (1st year - 4th year, orange arrows).}
 \label{fig:elementwise}
 \end{center}
\end{figure*} 
%%%%%%%%%%%%%%%%%%%%%%%%

We also present a real-world example of handwriting recognition \cite{Alimoglu1996digits} 
(\figref{fig:elementwise}{d} and SI, Section S3.2).
The same procedure reveals that some clusters of digits are correctly and consistently identified (``0''), while the error mostly results from incorrect grouping of other digit clusters (``9'', ``8'', and ``1'';  
\figref{fig:elementwise}{e}).
Element-wise frustration shows that there are some digits that cannot be consistently classified (``3'' and ``8'', 
\figref{fig:elementwise}{f}),
while some errors are regularly made (``1'' and ``9''). 
The extreme examples of these two types of error are shown in
\figref{fig:elementwise}{g}.

\subsection*{The convolution of meta-data in social networks}
%Finally, w
We now use our framework to explore the community structure of Facebook college friendship networks.
Previous research has suggested that friendship networks at major universities are organized into clusters which reflect the graduation year, dormitory, or student major \cite{Traud2011facebook, Traud2012facebook}.
However, the details of the organizing principles underlying this similarity are unknown.
Here we demonstrate and visualize how multiple attributes interact and contribute to community structure.

The Facebook friendship networks analyzed here were originally released as part of the the Facebook 100 data set \cite{Traud2011facebook, Traud2012facebook}.
This dataset contains a snapshot of all friendships at each of $100$ schools in the fall of $2005$.
Additionally, the data includes several categorical variables shared by the users on their individual pages: gender, class year, high school, major, and dormitory residence.
Here, we analyze the networks in two schools: the Oberlin (College $A$) and Rochester networks (College $B$).
For each school we took the largest connected component and uncovered clusterings using the Louvain method \cite{Blondel2008louvain}.
The categorical data for year, dorm and major were used to create three non-overlapping clusterings.
Every student with missing categorical data was placed into an individual singleton cluster.

Element-centric similarity reveals that school year closely captures the modular structure for most of the network, confirming previous results  \cite{Traud2011facebook, Traud2012facebook}.
However, our element-centric similarity further illustrates that this similarity is particularly high for the students in their 1st or 2nd years, and fails to capture the clustering structure of other students (\figref{fig:elementwise}{h,i} black arrows).
In these cases, the students' major gradually takes over the cohort-based connections (\figref{fig:elementwise}{h,i} red arrows).
This result, which has only become straight-forward through our framework, supports the intuition that network structure results from the convolution of multiple attributes \cite{Peel2017metadata}.

%%%%%%%%%%%%%%%%%%%%%%
% Figure for Functional Connectome Classification
\begin{figure*}
\begin{center}
	\includegraphics[width = 14cm]{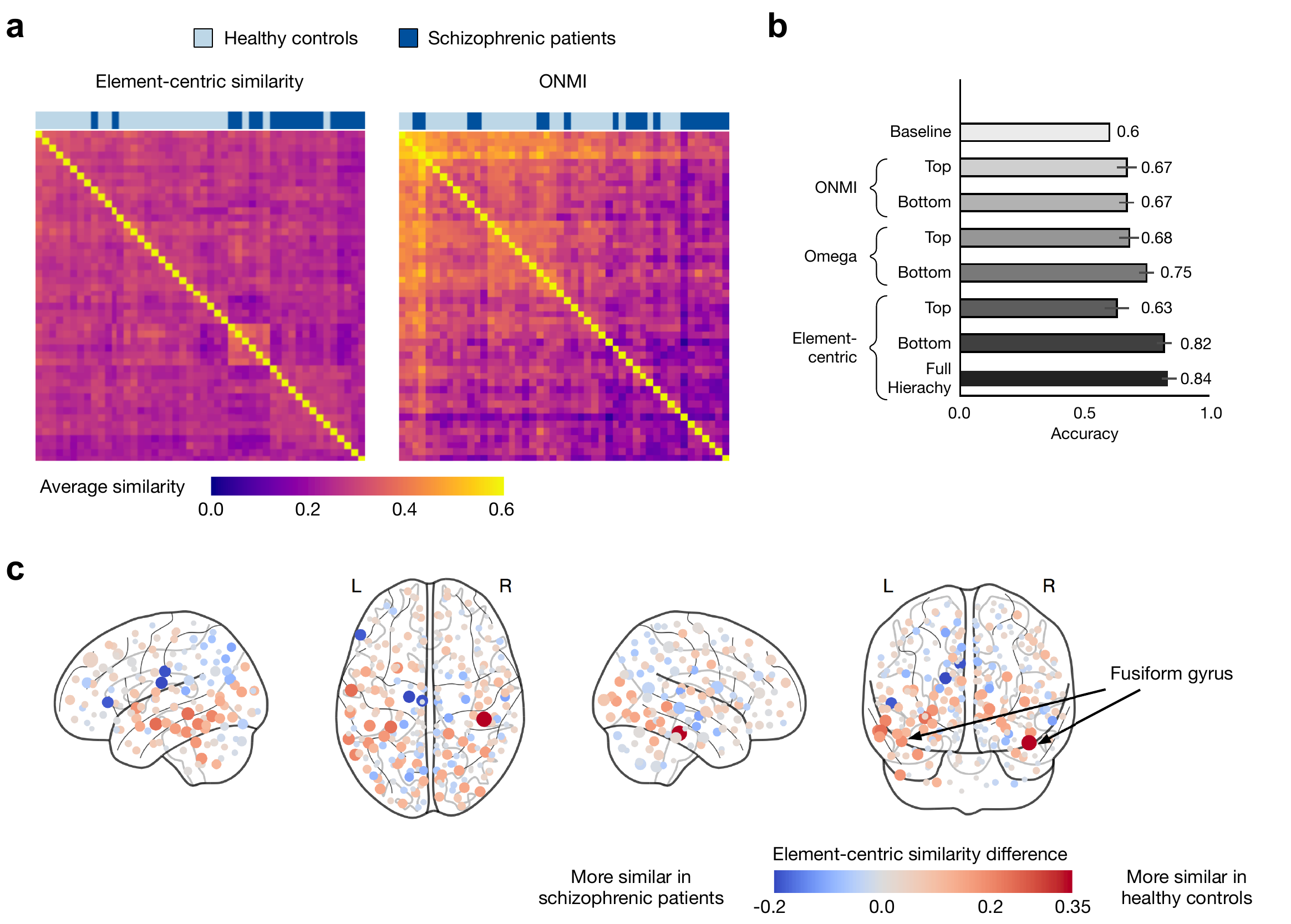}
	\caption{Our element-centric similarity better differentiates the overlapping and hierarchical community structure of functional brain networks in healthy and schizophrenic individuals. \textbf{a}, Hierarchical clustering of average pair-wise element-centric similarity using the entire OSLOM hierarchy closely reflects the true classification of participants as healthy (light blue) or schizophrenic (dark blue), while hierarchical clustering of the average pair-wise similarity using ONMI on the bottom level of the OSLOM hierarchy fails to uncover patient classification. \textbf{b}, Classification accuracy using different clustering similarity measures averaged over $100$ instances of $10$-fold cross-validation, error bars denote one standard deviation. \textbf{c}, The difference in element-centric similarity for each brain region when comparing amongst the healthy controls minus the similarity when comparing amongst the schizophrenic individuals; ROIs within the Fusiform gyrus are more consistently clustered in the healthy controls than in the schizophrenic individuals.}
 \label{fig:funcconn}
 \end{center}
\end{figure*}
%%%%%%%%%%%%%%%%%%%%%%%%

\subsection*{Element-centric comparisons of overlapping and hierarchical clustering in brain networks}

Finally, to further illustrate the utility of our element-centric similarity measure, to demonstrate its ability to capture meaningful differences in overlapping and hierarchical clustering structure by classifying schizophrenic individuals based on the community structure of resting-state fMRI brain networks.
There are several known distinctive and interpretable properties of resting-state fMRI brain networks in schizophrenia \cite{AlexanderBloch2012schizophreniadifferences,  Cheng2015centralityschizophrenia, Fornito2012schizophreniaconnectomics, Du2012schizophreniaaccuracy, Arbabshirani2013schizophreniaclassificaiton}.
Network communities, in particular, are hypothesized to capture functionally integrated modules in the brain that reflect key properties of schizophrenia \cite{AlexanderBloch2012schizophreniadifferences}.
Our goal for this example is not to introduce a superior classification of schizophrenic subjects, rather, upon controlling the clustering method and data set, we demonstrate that our measure can extract more useful information than the other state-of-the-art clustering comparison methods for overlapping clusterings (ONMI, Omega index).
We extract communities with overlapping and hierarchical structure using OSLOM community detection \cite{Lancichinetti2011oslom} from the functional brain networks of $48$ subjects ($29$ healthy controls and $19$ individuals diagnosed with schizophrenia) analyzed in a previous study \cite{Cheng2015centralityschizophrenia} (see SI, Section S3.3 for details).
The similarity between each pair of the subjects' hierarchical and overlapping clusterings was found using our element-centric similarity measure, producing a $48\times 48$ similarity matrix (\figref{fig:funcconn}{a}).

The subject-subject similarity matrix was then used in conjunction with a weighted k-nearest neighbors classifier to perform a binary classification of subjects as either schizophrenic or healthy controls.
Evaluated by a nested $10$-fold cross-validation procedure, our approach achieves an average accuracy of $84\%$, outperforming other measures (ONMI, the Omega index, \figref{fig:funcconn}{b}).
Note that, classification based on individual levels from the hierarchy does not perform as well as the method using the full hierarchy. 
Even when limited only to the overlapping clustering at the bottom of the OSLOM hierarchy, our element-centric clustering similarity outperforms both ONMI and the Omega Index.

Our element-centric clustering similarity measure also provides insights into which brain regions are consistently clustered within groups.
To find such group differences, we consider the element-centric similarity between all healthy controls, and the element-centric similarity between all schizophrenic patients.
As seen in \figref{fig:funcconn}{c}, the difference between the means of these two groups highlights several regions which are consistently clustered into similar functional modules in the healthy controls or schizophrenic patients.
In particular, regions of interest (ROIs) located in the Fusiform gyrus (Brodmann Area 37) were consistently clustered in the healthy controls but displayed great variability in cluster structure for the schizophrenic patients. 
This result is corroborated by the fact that the Fusiform gyrus has previously been associated with abnormal activation in schizophrenia during semantic tasks~\cite{kircher2001schizsentcomp, kuperberg2007schizsemantic}.

%%%%%%%%%%%%%%%%%%%%%%%%%%%%%%
% Discussion
%%%%%%%%%%%%%%%%%%%%%%%%%%%%%%
\section*{Summary and discussion}
In summary, we present an element-centric framework that intuitively unifies the comparison of disjoint, overlapping, and hierarchically structured clusterings.
We have presented that our element-centric similarity does not suffer from the common counter-intuitive biases of existing measures, and that it also provides insights into how clusterings differ at the level of individual elements.

Our framework suggests straight-forward extensions to more complex scenarios, such as soft or fuzzy clusterings, hierarchical clusterings specified by dendrograms with merge distance information, and hyper-graph similarity.
The framework also provides a measure of pair-wise similarity between elements, akin to the nodal association matrix of Bassett \emph{et al}.~\cite{Bassett2013robustcommunities}, and an element-wise clustering similarity which summarizes the difference in relationships induced by overlapping and hierarchically structured clusterings from the perspective of individual elements.
Both of these objects hold promise for use in clustering ensemble methods \cite{Monti2003consensus, Lancichinetti2012consensus}.

As clustering methods advance to uncover more nuanced and accurate organizational structure of complex systems, so too should clustering similarity measures facilitate meaningful comparisons of these organizations.
The element-centric framework proposed here provides an intuitive quantification of clustering similarity that holds great promise for uncovering the relationships amongst all types of clusters, such as network communities, ontogenies, and dendrograms.
The ubiquity of clustering in all areas of science suggests extensive potential impact of our framework.

\section*{Acknowledgements}
We would like to thank Dae-Jin Kim for assistance with the interpretation of the schizophrenia classifications, and Hu Cheng for assistance processing the fMRI timeseries.  We thank Randall D. Beer, Luis M. Rocha, Filippo Radicchi, Sune Lehmann, Olaf Sporns, Alessio Cardillo, and Artemy Kolchensiky for helpful discussions.  Supported in part by National Institute for Mental Health Grant 2R01MH074983 to WPH.  YYA thanks Microsoft Research for support through a Microsoft Research Faculty Fellowship. We thank the anonymous reviewers for their insightful comments.

\section*{Additional Information}

\subsection*{Author Contributions}
AJG and YYA developed the method, AJG and IBW performed the analysis, WPH contributed the fMRI data, AJG, IBW, WPH, and YYA participated in interpreting results, AJG and YYA wrote the manuscript. All authors reviewed and edited the manuscript.

\subsection*{Competing Interests}
The authors declare no competing interests.

\subsection*{Data Availability}
All data used in this work is available upon request.  A full implementation of the element-centric similarity measure is available in the open-source package: CluSim~\cite{Gates2018clusim}.

\pagebreak
\clearpage

{\huge\bfseries Supplemental Information}\\[0.25in]

{\LARGE Element-centric clustering comparison unifies overlaps\\ and hierarchy}

% Author and supervisor
Alexander J.\ Gates$^{1}$, Ian B. Wood$^{2,3}$, William P. Hetrick$^{4}$, and Yong-Yeol Ahn$^{2,3,5}$

$^{1}$Department of Physics, Northeastern University. Boston, MA

$^{2}$Department of Informatics, Indiana University. Bloomington, IN

$^{3}$Center for Complex Networks and Systems Research, Indiana University. Bloomington, IN

$^{4}$Department of Psychological and Brain Sciences, Indiana University. Bloomington, IN

$^{5}$Program in Cognitive Science, Indiana University. Bloomington, IN

%Center for Complex Network Research, Northeastern University, Boston, Massachusetts 02115, USA

\tableofcontents

\pagebreak

\beginsupplement

\section{Clusterings}
\label{sec:clusterings}
%%%%%%%%%%%%%%%
Throughout this work, we focus on the grouping of elements (i.e. data points or vertices) into clusters (the groups).
The set of clusters is called a \emph{clustering}.
Specifically, given a set of $N$ distinct elements $V = \{v_{1},\ldots,v_{N}\}$, a clustering is a set $\mathcal{C} = \{C_1,\ldots, C_{K_{\mathcal{C}}}\}$ of $K_{\mathcal{C}}$ non-empty subsets of $V$ such that every element $v_i$ in $V$ is in at least one cluster $C_\beta$: $\forall v_i\in V$ $\exists C_\beta$ s.t.\ $v_i\in C_\beta$.

We consider three classes of clusterings.
A \emph{partition}, or disjoint clustering, is a clustering in which all elements are members of one, and only one, cluster.
An \emph{overlapping} clustering allows elements to be members of multiple clusters.
\emph{Hierarchical} clusterings capture the nested organization of clusters at different scales and are accompanied by a directed acyclic graph (or dendrogram) showing the hierarchical relationships between clusters.

\section{Existing measures of clustering similarity}
\label{sec:clusteringsim}
%%%%%%%%%%%%%%%

Here, we focus on ten of the most prominent measures from the clustering literature: the Rand index, the adjusted Rand index, the Omega index, the Jaccard index, the F measure, the Fowlkes Mallows index, percentage matching (PM), normalized mutual information (NMI), overlapping normalized mutual information (ONMI), variation of information (VI).
All of these measures are implemented in the CluSim python package\cite{Gates2018clusim}.

\subsection{Rand Index}
\label{sec:rand}
The Rand index \cite{Rand1971randindex} counts the number of element pairs which are either members of the same cluster, or members of different clusters in both clusterings.
The most common formulation of the Rand index focuses on the following four sets of the $\binom{N}{2}$ element pairs: 
$N_{11}$ the number of element pairs which are grouped in the same cluster in both clusterings, 
$N_{10}$ the number of element pairs which are grouped in the same cluster by $\mathcal{A}$ but in different clusters by $\mathcal{B}$, 
$N_{01}$ the number of element pairs which are grouped in the same cluster by $\mathcal{B}$ but in different clusters by $\mathcal{A}$, 
and $N_{00}$ the number of element pairs which are grouped in different clusters by both $\mathcal{A}$ and $\mathcal{B}$. 
Intuitively, $N_{11}$ and $N_{00}$ are indicators of the agreement between the two clusterings, while $N_{10}$ and $N_{01}$ reflect the disagreement between the clusterings.

\begin{table}[b!]
 	\begin{center}
    	\begin{tabular}{c|c c c c|c}
	$\mathcal{A} / \mathcal{B}$ & $B_1$ & $B_2$ & $\hdots$ & $B_{K_{\mathcal{B}}}$ & Sums \\ \hline
	$A_1$ & $n_{11}$ & $n_{12}$ & $\hdots$ & $n_{1K_{\mathcal{B}}}$ & $a_1$ \\
	$A_2$ & $n_{21}$ & $n_{22}$ & $\hdots$ & $n_{2K_{\mathcal{B}}}$ & $a_2$ \\
	$\vdots$ & $\vdots$ & $\vdots$ & $\ddots$ & $\vdots$ &$\vdots$ \\
	$A_{K_{\mathcal{A}}}$ & $n_{K_{\mathcal{A}}1}$ & $n_{K_{\mathcal{A}}2}$ & \ldots & $n_{K_{\mathcal{A}}K_{\mathcal{B}}}$ & $a_{K_{\mathcal{A}}}$ \\ \hline
	Sums & $b_{1}$ & $b_{2}$ & $\hdots$ & $b_{K_{\mathcal{B}}}$ & $\sum_{ij}n_{ij} = N$
	\end{tabular}
	\end{center}
	\caption{The contingency table $\mathcal{T}$ for two clusterings $\mathcal{A} = \{A_1, \ldots, A_{K_{\mathcal{A}}}\}$ and $\mathcal{B} = \{B_1,\ldots, B_{K_{\mathcal{B}}}\}$ of $N$ elements, where $n_{ij} = |A_i\cap B_j|$ are the number of elements in both cluster $A_i\in\mathcal{A}$ and cluster $B_j\in\mathcal{B}$.}
	\label{tbl:cont}
\end{table}

The aforementioned pair counts are identified from the contingency table $\mathcal{T}$ between two clusterings, shown in Table \ref{tbl:cont}, by the following set of equations:
\begin{linenomath*}
\begin{align}
	N_{11} &= \sum_{k,m = 1}^{K_\mathcal{A},K_\mathcal{B}}\binom{n_{km}}{2} = \frac{1}{2}\left(\sum_{k,m = 1}^{K_\mathcal{A},K_\mathcal{B}}n_{km}^2 - N\right)  \\
	N_{10} &= \sum_{k=1}^{K_\mathcal{A}} \binom{a_k}{2} -N_{11} = \frac{1}{2}\left(\sum_{k=1}^{K_\mathcal{A}} a_k^2 - \sum_{k,m = 1}^{K_\mathcal{A},K_\mathcal{B}}n_{km}^2\right) \nonumber \\
	N_{01} &= \sum_{m=1}^{K_\mathcal{B}} \binom{b_m}{2} -N_{11} = \frac{1}{2}\left(\sum_{m=1}^{K_\mathcal{B}} b_m^2 - \sum_{k,m = 1}^{K_\mathcal{A},K_\mathcal{B}}n_{km}^2\right) \nonumber \\
	N_{00} &= \binom{N}{2} - N_{11} - N_{10} - N_{01}. \nonumber
\end{align}
\end{linenomath*}
The Rand index between clusterings $\mathcal{A}$ and $\mathcal{B}$, $RI(\mathcal{A},\mathcal{B})$ is then given by the function:
\begin{linenomath*}
\begin{align}
	RI(\bm{A},\bm{B}) &= \frac{N_{11} + N_{00}}{\binom{N}{2}}.
	\label{eq:rand}
\end{align}
\end{linenomath*}
It lies between $0$ and $1$, where $1$ indicates the clusterings are identical and $0$ occurs for clusters which do not share a single pair of elements (this only happens when one clustering is the full set of elements and the other clustering groups each element into its own singleton cluster).
As the number of elements being clustered becomes large, the measure becomes dominated by the number of pairs which were classified into different clusters ($N_{00}$), resulting in decreased sensitivity to co-occurring element pairs \cite{Fowlkes1983hierarchicalcompare}.

\subsection{Adjusted Rand index (ARI)}
A popular extension of the Rand index, called the adjusted Rand index (ARI), considers the average of the measure in the context of the permutation model for random clusterings \cite{Hubert1985adjrand, Albatineh2006corrchance, Gates2017impact}.
In the permutation model the number and size of clusters within a clustering are fixed; all random clusterings are generated by shuffling the elements between the fixed clusters.
The expectation of the Rand index with respect to the permutation model follows from the fact that the entries in Table \ref{tbl:cont} follow a generalized hypergeometric distribution.
Taking $Q^\mathcal{A} = \sum_{k=1}^{K_{\mathcal{A}}}\binom{a_k}{2}$ and $Q^\mathcal{B} = \sum_{m=1}^{K_{\mathcal{B}}}\binom{b_m}{2}$, the expectation $\mathbb{E}_{perm}[RI(\mathcal{A},\mathcal{B})]$ of the Rand index with respect to the permutation model for the cluster size sequences of clusterings $\mathcal{A}$ and $\mathcal{B}$ is given by:
\begin{linenomath*}
\begin{align}
	\mathbb{E}_{perm}[RI(\mathcal{A},\mathcal{B})] = \frac{Q^\mathcal{A}Q^\mathcal{B} -  \binom{N}{2}\Big(Q^\mathcal{A} + Q^\mathcal{B}\Big) + \binom{N}{2}^2}{\binom{N}{2}^2}
	\label{eq:hyperexprand}
\end{align}
\end{linenomath*}
(see Fowlkes and Mallows \cite{Fowlkes1983hierarchicalcompare}, Hubert and Arabie \cite{Hubert1985adjrand}, or Albatineh and Niewiadomska-Bugaj \cite{Albatineh2006corrchance} for the full derivation).
Finally, the ARI between clusterings $\mathcal{A}$ and $\mathcal{B}$ is given by:
\begin{linenomath*}
\begin{align}
	\text{ARI}(\mathcal{A},\mathcal{B}) &=  \frac{R(\mathcal{A},\mathcal{B}) - \mathbb{E}_{perm}[RI(\mathcal{A},\mathcal{B})]}{1 - \mathbb{E}_{perm}[RI(\mathcal{A},\mathcal{B})]} 
\end{align}
\end{linenomath*}

\subsection{Omega index}
The Omega index extends the adjusted Rand index to compare overlapping clusterings \cite{Collins1988omega}.  
To formulate the extension, notice that in the presence of overlaps, element pairs can repeatedly occur within the same cluster.  
We consider $t_j(\mathcal{A})$ the set of node pairs which co-occur exactly $j$ times in clustering $\mathcal{A}$.
The unadjusted Omega index between two overlapping clusterings is then:
\begin{linenomath*}
\begin{equation}
    \omega_{u}(\mathcal{A},\mathcal{B}) = \frac{1}{\binom{N}{2}}\sum_{j}|t(\mathcal{A})\cap t(\mathcal{B})|,
\end{equation}
\end{linenomath*}
while the expectation of this measure with respect to the permutation model on the number of element pair overlaps is:
\begin{linenomath*}
\begin{equation}
    \mathbb{E}_{perm}[\omega_{u}(\mathcal{A},\mathcal{B})] = \frac{1}{\binom{N}{2}^2}\sum_{j}|t(\mathcal{A})|\cdot|t(\mathcal{B})|
\end{equation}
\end{linenomath*}
Finally, the Omega index between two overlapping partitions is given by:
\begin{linenomath*}
\begin{equation}
    \Omega(\mathcal{A},\mathcal{B}) = \frac{\omega_{u}(\mathcal{A},\mathcal{B}) - \mathbb{E}_{perm}[\omega_{u}(\mathcal{A},\mathcal{B})]}{1 - \mathbb{E}_{perm}[\omega_{u}(\mathcal{A},\mathcal{B})]}.
\end{equation}
\end{linenomath*}
Note that for partitions the Omega index is equivalent to the adjusted Rand index.

\subsection{Jaccard index}
Another popular clustering similarity measure which utilizes pair-wise co-occurrence between the elements is the Jaccard index or Jaccard similarity coefficient \cite{BenHur2002stability}.
Originally proposed to compare regional floras \cite{Jaccard1912flora}, the Jaccard index is a similarity measure for finite sets.
It is defined as the number of element pairs which are grouped in the same cluster in both clusterings divided by the number of element pairs which are grouped in the cluster in at least one of the clusterings.  
Thus, it ignores the number of element pairs that are grouped into different clusters by both clusterings.
One minus the Jaccard index is a metric on the collection of finite sets \cite{Marczewski1958setdistance}.
Using the above notation from the contingency table Table \ref{tbl:cont}, the Jaccard index between clusterings $\mathcal{A}$ and $\mathcal{B}$ takes the form:
\begin{linenomath*}
\begin{align}
    \text{J}(\mathcal{A}, \mathcal{B}) &= \frac{N_{11}}{N_{11} + N_{10} + N_{01}}
\end{align}
\end{linenomath*}

\subsection{F measure}
The F measure has a long history of use in clustering validation, natural language processing, information retrieval, and machine learning.
It is based off of two asymmetric measures (sometimes called Dice's asymmetric coefficients), that count the proportion of element pairs which are correctly co-assigned to the same cluster in one of the clusterings using the other clustering as a baseline.
When one of these clusterings is considered to be a ground-truth clustering, these asymmetric measures are known as \emph{precision} and \emph{recall}.
The F measure is the harmonic mean of the precision and recall.
Specifically, the F measure between clusterings $\mathcal{A}$ and $\mathcal{B}$ is given by:
\begin{linenomath*}
\begin{align}
    F(\mathcal{A}, \mathcal{B}) &= \frac{2N_{11}}{2N_{11} + N_{10} + N_{01}}
\end{align}
\end{linenomath*}
The F measure $F$ and Jaccard index $J$ are related by $J = F/(2-F)$.

\subsection{Fowlkes-Mallows index}
The Fowlkes-Mallows index was first introduced to facilitate the comparison of hierarchical dendrograms \cite{Fowlkes1983hierarchicalcompare}.
The idea is to cut the dendrogram at each merger and compare the induced flat clusterings. 
Like the previous five measures, the Fowlkes-Mallows index is based on counting the pair-wise co-occurrence between the elements in the two clusterings:  
\begin{linenomath*}
\begin{equation}
    \text{FM}(\mathcal{A}, \mathcal{B}) = \frac{N_{11}}{\sqrt{(N_{11} + N_{10}) (N_{11} + N_{01})}}.
\end{equation}
\end{linenomath*}
Applying this index to each cut $k$ of two dendrograms produces a curve of comparisons between two clusterings each with $k$ clusters.

%{\color{red}
\subsection{Percentage Matching}
The Percentage Matching is based on the idea that each cluster should be compared to only one other cluster, its ``best match'' \cite{Meila2001experimental}.
Specifically, let $K_{\text{min}} = \min(K_{\mathcal{A}}, K_{\mathcal{B}})$, then the percentage matching index is defined using the contingency table:
\begin{linenomath*}
\begin{equation}
    \text{PM}(\mathcal{A}, \mathcal{B}) = 1-\frac{1}{N}\sum_{k=1}^{K_{\text{min}}}\max_{\pi}n_{k,\pi}.
\end{equation}
\end{linenomath*}
where the notation $\max_{\pi}$ denotes finding the cluster $\pi$ with the largest overlap to cluster $k$.
The percentage matching is equal to one minus the Purity Index, another common measure of the distance between clusterings.
%}

\subsection{Normalized mutual information (NMI)}
\label{sec:nmi}
Another family of approaches for finding the similarity of two cluster coverings is based on the amount of information in each covering and the amount of information one covering contains about the other.
These quantities can also be calculated from the contingency Table \ref{tbl:cont}.  The entropy $H$ of a clustering $\mathcal{A}$ is given by 
\begin{linenomath*}
\begin{equation}
	H(\mathcal{A}) = - \sum_{k=1}^{K_{\mathcal{A}}}\frac{a_k}{N}\log\frac{a_k}{N}
\end{equation}
\end{linenomath*}
and, using the entries $n_{km}$ from the contingency table \ref{tbl:cont}, the joint entropy between two clusterings $\mathcal{A}$ and $\mathcal{B}$ is 
\begin{linenomath*}
\begin{equation}
    	H(\mathcal{A}, \mathcal{B}) = - \sum_{k,m = 1}^{K_{\mathcal{A}}, K_{\mathcal{B}}}\frac{n_{km}}{N}\log\frac{n_{km}}{N}
\end{equation}
\end{linenomath*}
Thus, the mutual information between two clusterings is given by:
\begin{linenomath*}
\begin{align}
	MI(\mathcal{A}, \mathcal{B}) &= H(\mathcal{A}) + H(\mathcal{B}) - H(\mathcal{A}, \mathcal{B}) \nonumber \\
			&= \sum_{k,m = 1}^{K_{\mathcal{A}}, K_{\mathcal{B}}}\frac{n_{km}}{N}\log\frac{n_{km} N}{a_k b_m}.
\end{align}
\end{linenomath*}
The mutual information can be interpreted as an inverse measure of independence between the clusterings, or a measure of the amount of information each clustering has about the other.
As it can vary in the range $[0,\min\{H(\mathcal{A}),H(\mathcal{B})\}]$, to facilitate comparisons, it is desirable to normalize it to the range $[0,1]$.  There are at least six proposals in the literature for this upper bound, each with different advantages and drawbacks; 
\begin{linenomath*}
\begin{align}
	\label{eq:mimax}
	\min\{H(\mathcal{A}),H(\mathcal{B})\} \leq \sqrt{H(\mathcal{A})H(\mathcal{B})} \leq \frac{H(\mathcal{A}) + H(\mathcal{B})}{2}  \\ 
	 \leq \max\{H(\mathcal{A}),H(\mathcal{B})\} \leq \max\{\log K_{\mathcal{A}}, \log K_{\mathcal{B}}\} \leq \log N. \nonumber
\end{align}
\end{linenomath*}
The resulting measures are all known as normalized mutual information (NMI).  Here, we always use the average of the two clustering entropies $ \frac{1}{2}\left(H(\mathcal{A}) + H(\mathcal{B})\right)$.  Although some results have been shown to depend on the normalization term used for NMI, \figref{fig:nmicompare} demonstrates that NMI behaves un-intuitively regardless of the normalization term.

Due to the known bias of NMI towards clusterings with more clusters, several modifications have been proposed.
The NMI can be adjusted for chance according to an appropriate random model \cite{Vinh2009nmicorrection, Gates2017impact}, but this induces the problem of selecting a random model for the clusterings, and does not remove the issue of selecting a normalization term.
Alternatively, the NMI can be re-scaled by an exponential factor reflecting the difference in number of clusters between the two clusterings, but this scaling factor forces the researcher to prioritize one clustering as the 'ground-truth' and breaks the symmetry of the original measure \cite{Amelio2015nmifair}.
%

%%%%%%%%%%%%%%%%%%%%%%
% Figure with CSTG
\begin{figure*}
	\includegraphics[width = 0.9\textwidth]{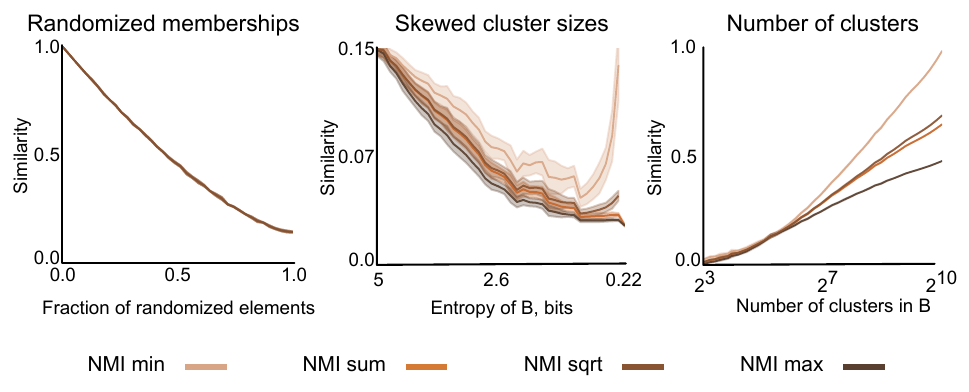}
	\caption{NMI's bias towards the number of clusters is independent of normalization term.  The three scenarios from the main text, for different normalization terms of NMI: the minimum of cluster entropies (min), the average of the cluster entropies (sum), the geometric mean of the cluster entropies (sqrt), and the maximum of the cluster entropies (max).  See Section \ref{sec:nmi} for the measure details. } 
 \label{fig:nmicompare}
\end{figure*}
%%%%%%%%%%%%%%%%%%%%%%%%

\subsection{Overlapping NMI (ONMI)}	
The NMI has been modified to account for clusterings with overlapping clusters \cite{Lancichinetti2009detectingcommunity}.
Consider a clustering $\mathcal{A}$ with $K_{\mathcal{A}}$ possibly overlapping clusters $A_1, \hdots, A_{K_{\mathcal{A}}}$.
For each cluster $A_k$, we can consider a binary random variable $X_k$ which represents the probability that a randomly selected node is a member of that cluster with distribution
\begin{linenomath*}
\begin{equation}	
	P(X_k = 1) = \frac{a_k}{ N}, \hspace{8pt} P(X_k = 0) = 1 - \frac{a_k}{ N}
\end{equation}
\end{linenomath*}
The same holds for a second clustering $\mathcal{B}$ with $K_{\mathcal{B}}$ possibly overlapping clusters $B_1, \hdots, B_{K_{\mathcal{B}}}$ and random variables $Y_m$.
We can then define the joint probability distribution $P(X_k, Y_m)$:
\begin{linenomath*}
\begin{align}
	P(X_k = 1, Y_m = 1) = \frac{n_{km} }{ N} \nonumber \\
	P(X_k = 0, Y_m = 0) = 1 - \frac{n_{km} }{ N} \\
	P(X_k = 1, Y_m = 0) = \frac{a_k - n_{km}}{N} \nonumber \\
	P(X_k = 0, Y_m = 1) = \frac{b_m - n_{km}}{N} \nonumber
\end{align}
\end{linenomath*}
Given a particular cluster $A_k\in\mathcal{A}$, the amount of information it has about another cluster $B_m\in\mathcal{B}$ is found by the conditional entropy
\begin{linenomath*}
\begin{equation}
	H(X_k|Y_m) = H(X_k,Y_m) - H(Y_m).
\end{equation}
\end{linenomath*}
In the case of overlapping clusters, there are many possible candidates for the best match between two clusters.  The best match is the one with the minimal conditional entropy.  Thus, the conditional entropy of $X_k$ with respect to all of the clusters in $\mathcal{B}$ is 
\begin{linenomath*}
\begin{equation}
	H(X_k|\bm{Y}) = \min_{m \in\{ 1,\hdots,M\}} H(X_k|Y_m).
\end{equation}
\end{linenomath*}
However, in minimizing the entropy it may be the case that the optimal $B_m^*$ is the complement of $A_k$, thus we have to add the following constraint to insure the above minimization identities the best matching cluster:
%4
\begin{linenomath*}
\begin{equation}
	h[P(1,1)] + h[P(0,0)] > h[P(0,1)] + h[P(1,0)].
\end{equation}
\end{linenomath*}
This entropy is normalized by the entropy of $X_k$ and averaged over all $X_k$ to give the normalized conditional entropy of $\bm{X}$ with respect to $\bm{Y}$
\begin{linenomath*}
\begin{equation}
	H(\bm{X}|\bm{Y})_{\text{norm}} = \frac{1}{K}\sum_{k =1}^K \frac{H(X_k|\bm{Y})}{H(X_k)}.
\end{equation}
\end{linenomath*}
Finally, the overlapping normalized mutual information ONMI is given by
\begin{linenomath*}
\begin{equation}
	ONMI(\mathcal{A},\mathcal{B}) = 1 - \frac{1}{2}[H(\bm{X}|\bm{Y})_{\text{norm}} + H(\bm{Y}|\bm{X})_{\text{norm}}].
\end{equation}
\end{linenomath*}

It is interesting to note that when $\mathcal{A}$ and $\mathcal{B}$ are partitions, the $NMI(\mathcal{A},\mathcal{B})$ and $ONMI(\mathcal{A},\mathcal{B})$ do not necessarily agree.
Although there have been several attempts to reformulate ONMI so that it agrees with NMI, the above formulation is pervasive in the literature \cite{Esquivel2012onmi, Xie2013overlappingcommunities, Hric2014groundtruth}.

\subsection{Variation of Information VI}
\label{sec:vi}
Another popular clustering comparison measure based on information theory is the Variation of Information (VI).
Unlike the similarity measures discussed above, the VI is a metric on the lattice of partitions \cite{Meila2003comparingvi}.
Thus, it is a measure of distance between clusterings instead of a similarity between the clusterings; it attains its minimum at $0$ when the clusterings are identical, and attains positive values for clusterings which differ.
Using the entropy and mutual information between clusterings defined in Section \ref{sec:nmi}, the VI is given by:
\begin{linenomath*}
\begin{align}
	VI(\mathcal{A}, \mathcal{B}) &= H(\mathcal{A}) + H(\mathcal{B}) - 2MI(\mathcal{A}, \mathcal{B}) \nonumber \\
			&= 2H(\mathcal{A}, \mathcal{B}) - H(\mathcal{A}) - H(\mathcal{B}).
\end{align}
\end{linenomath*}

Since the VI is a distance measure, the intuitive behavior is opposite that presented for the similarity measures discussed in this paper, and presented in the main text, Figure 2.
None-the-less, we can demonstrate that the VI suffers from unintuitive behavior in two scenarios: the skewed cluster sizes and the number of clusters (\figref{fig:vibias}).

%%%%%%%%%%%%%%%%%%%%%%
% Figure with CSTG
\begin{figure*}
	\includegraphics[width = 0.9\textwidth]{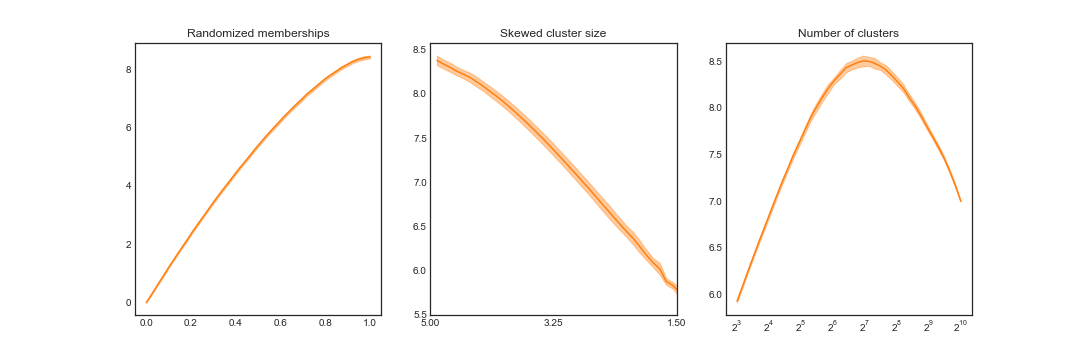}
	\caption{VI unintuitive behavior as the cluster sizes become more skewed and as the number of clusters is increased.  Note that because the VI is a distance measure, the intuitive behavior is opposite that presented for similarity measures.}
 \label{fig:vibias}
\end{figure*}
%%%%%%%%%%%%%%%%%%%%%%%%

%{\color{red}
\subsection{Information Theoretic Intuition}
A second intuition that could be used to evaluate clustering similarity measures is based on concepts drawn from information theory.
Under this intuition, the appropriate question to ask is: ``Given a random element, how much uncertainty remains about its membership in Clustering $\mathcal{B}$ if I know its membership in Clustering $\mathcal{A}$?''
The clusters are now considered as an alphabet and the contingency table is considered as a discrete probability distribution over this alphabet.
For example, the variation of information considers the difference in conditional entropies reflecting the amount of information we loose about the original cluster assignment, and the amount of information we have to gain to recover the new cluster assignment when going from one clustering to the other\cite{Meila2003comparingvi}.
The resulting intuition suggests that two clusterings are similar if one doesn't loose much information (presence of equally sized clusters) or one doesn't have to gain much information (presence of very small clusters).
Consequently, in \figref{fig:vibias}, we notice that the VI decreases (more similar) as the cluster entropy decreases, and displays a parabolic shape (more similar, to less similar, to more similar) as the number of clusters approaches the number of elements.

Our main objection to the information theoretic intuition is that it tends to suggest measures cannot differentiate the influence of alphabet size (here, number of clusters) from the distribution of alphabet usage (here, sizes of the clusters).
Furthermore, the primary justification for these measures is typically stated with respect to the alignment to the lattice of partitions \cite{Meila2005comparingclusteringsaxiom}, yet, it is not immediately clear if the lattice of partitions is the appropriate space to compare clustering similarity measures since many applications do not align to the lattice (i.e.\ evaluation of k-means clustering fixes the number of clusters).
%}

\section{Datasets}
\subsection{Point clusters}
\label{sec:pointclusters}
$5,000$ points were random formed into clusters in an algorithm akin to the process for constructing benchmark graphs \cite{Lancichinetti2009comparecommunity}.
Cluster sizes were randomly drawn from a powerlaw distribution with a minimum cluster size of $10$, a maximum cluster size of $1000$, and an exponent of $1.0$.
The center of those clusters was uniformly selected from points in a $40\times40$ box.  
The standard deviation (or spread) of each cluster was also drawn from a powerlaw distribution with a minimum of $0.2$, a maximum of $2.0$, and an exponent of $1.0$.
Next, the type of each cluster was uniformly selected from four options.  The first option is the 2-D Gaussian blob with mean given by the cluster center and standard deviation given by the cluster standard deviation.
The second option is the 2-D Anisotropic blob with a mean given by the cluster center, standard deviation given by the cluster standard deviation, and transformation given by the rotational matrix:
\begin{linenomath*}
\begin{equation}
    \begin{bmatrix}
        a\cos(\theta) & -a\sin(\theta) \\
        b\sin(\theta) & b\cos(\theta)
    \end{bmatrix},
    \label{eq:rotation}
\end{equation}
\end{linenomath*}
where $a,b$ randomly drawn from the unit interval and $\theta$ was randomly drawn from the range $[0,\pi]$.
The third option is the circle centered at the cluster center with radius given by the cluster standard deviation; the points were uniformly spread along the circle and Gaussian noise with mean $0$ and standard deviation $0.2$ was added to all points.
The forth option is the spiral with points uniformly spread in the range $[0,10]$, converted to circular coordinates by $(x,y) \rightarrow (\sigma\sqrt{x}\cos(x), \sigma\sqrt{y}\cos(y))$, where $\sigma$ is the cluster standard deviation, randomly rotated by the rotation matrix of equation $\eqref{eq:rotation}$ with $a=b=1$ and $\theta$ randomly drawn from the range $[0,\pi]$, and Gaussian noise with mean $0$ and standard deviation $0.2$ was added to all points.

The sci-kit learn \cite{scikitlearn} implementation of $K$-means clustering was initialized with $K=19$ clusters and random initial centroids.
The identification method was then run from $100$ random centroid initializations. 
Clustering agreement was calculated by comparing all $100$ uncovered clusterings with the ground-truth clustering using the element-wise similarity vector was found for each comparison and then averaged over the uncovered clusterings.
Clustering frustration was calculated from all pair-wise comparisons between the $100$ uncovered clusterings using the element-wise similarity vector was found for each comparison and then averaged over each comparison.

\subsection{Handwriting digits}
\label{sec:handwriting}
The digits data set, originally assembled by Alimoglu and Alpaydin \cite{Alimoglu1996digits}, is bundled with the  \cite{scikitlearn} source code.
It consists of $1797$ images of $8 \times 8$ gray level pixels for handwritten digits distributed across $10$ clusters corresponding to the true digit.
To provide a visualization, the data was projected to 2-d using the t-Distributed Stochastic Neighbor Embedding (t-SNE) dimensionality reduction method \cite{vanderMaaten2008tsne} initialized from the pca decomposition.

The sci-kit learn \cite{scikitlearn} implementation of $K$-means clustering was initialized with $K=10$ clusters and random initial centroids.
The identification method was then run from $100$ random centroid initializations. 
Clustering agreement was calculated by comparing all $100$ uncovered clusterings with the ground-truth clustering using the element-wise similarity vector was found for each comparison and then averaged over the uncovered clusterings.
Clustering frustration was calculated from all pair-wise comparisons between the $100$ uncovered clusterings using the element-wise similarity vector was found for each comparison and then averaged over each comparison.

\subsection{Brain networks}
\label{sec:brainnetworks}

The dataset used here was originally analyzed in Cheng et al.\ \cite{Cheng2015centralityschizophrenia}; please refer to that work for specific details of the data acquisition and pre-processing, here we only provide a brief overview.

Data was acquired from $19$ individuals diagnosed with schizophrenia (6 female, mean age $33.1 \pm 10.9$ years) and $29$ healthy controls (15 female, mean age $28.1 \pm 8.4$ years). 
Diagnosis of schizophrenia was based on the Structured Clinical Interview for the DSM-IV Axis I Disorders (SCID-I) \cite{First2005schizophrenia} and medical chart review.  
All subjects were scanned on a Siemens TIM Trio $3$ T MRI scanner using a $32$-channel head coil. The high  anatomical scan had a resolution of $1$ mm$^3$.  
A total of $200$ volumes of resting state fMRI data were acquired with EPI sequences for $8$ min and $20$s. 
During the resting state fMRI scan, the subjects were at rest with eyes closed and instructed not to think of anything in particular.  
All functional data were motion corrected in FSL.

In conjunction with the anatomical image, the functional images were parcellated using a parcellation scheme proposed by Shen et al.\ \cite{Shen2013parcellation}. 
This parcellation divides the cerebral cortex into $278$ regions of interest (ROIs), and was derived from resting state functional data of the healthy subjects by maximizing functional homogeneity within each ROI. 
After regressing out head motion, the time signal was band-pass filtered between $0.01-0.10$ Hz and the time courses were extracted from the $278$ brain ROIs as the average over voxels.

The functional network was computed from the wavelet coherence between all pair-wise combinations of ROIs, giving rise to a square symmetric matrix ($278\times278$).  
The resulting functional connectivity matrix has only positive edges.  
In order to identify a backbone network structure, the multiscale network backbone \cite{Serrano2009multiscalebackbone} was extracted using an alpha of $\alpha = 0.2$.  
Technically, the multiscale backbone is a directed network, however, since our original graph was undirected, we convert the mutliscale backbone back into an undirected network.  
The network was not corrected to insure a single connected component.

Overlapping and hierarchically structured clusterings were derived using Order Statistics Local Optimization Method (OSLOM) network community detection \cite{Lancichinetti2011oslom} with the following parameters: weighted, undirected edges, $p=0.1$, $100$ runs for the detection at the bottom of the hierarchy and $1000$ runs for the detection at the top of the hierarchy.  
All singlet communities were kept in the clusterings.  
Due to the variability in clustering structure between runs of the algorithm, $10$ clusterings were extracted for each patient.

The subject similarity matrix was then constructed as follows.  
The similarity of each diagonal entry is $1.0$.  Each off-diagonal entry in the ($48\times48$) subject similarity matrix is the average element-centric similarity of all comparisons $10\times10 = 100$ between the $10$ OSLOM communities uncovered for each subject.  
For all comparisons, we set $\alpha = 0.9$ and $r=8.0$.  Our choice of the scaling parameter, $r=8.0$, was grounded in the explorations of synthetic binary hierarchies of equivalent height.  
The dis-similarity matrix is one minus the similarity matrix.  Six additional matrices were found by using the community structure found by slicing each OSLOM community dendrogram and retaining only the bottom or top communities and performing all pair-wise comparisons with either our element-centric similarity measure, ONMI or the Omega index.  Note that we use only these three measure of similarity because the communities contain many overlapping structures.

Given a dis-similarity matrix, a distance weighted k-Nearest Neighbors (kNN) classifier was trained using nested and stratified $10$-fold validation \cite{Hastie2001elements}.  
Specifically, the data was randomly split into $10$ groups such that the proportions of each class were kept relatively equal in each group.  
Each group in turn was then used as the testing set, while the other $9$ groups formed the training set.  For each training set, we first find the best $k$ for the kNN classifier using a grid search for $k$ between $1$ and $15$ and another stratified $10$-fold validation.  
The classifier was then retrained on the entire training set for the specified $k$.  
Finally, the accuracy of the trained classifier was found on the testing set.  In the paper, we report the average accuracy identified in $100$ random initializations of the nested $10$-fold validation technique \cite{Stone1974cross, Rao2008dangers}.

\end{document}